# Enhancing Marine Debris Acoustic Monitoring by Optical Flow-Based Motion Vector Analysis

Xiaoteng Zhou
*Graduate School of Frontier Sciences*
*The University of Tokyo*

Katsunori Mizuno
*Graduate School of Frontier Sciences*
*The University of Tokyo*

*Abstract*—With the development of coastal construction, a large amount of human-generated waste, particularly plastic debris, is continuously entering the ocean, posing a severe threat to marine ecosystems. The key to effectively addressing plastic pollution lies in the ability to autonomously monitor such debris. Currently, marine debris monitoring primarily relies on optical sensors, but these methods are limited in their applicability to underwater and seafloor areas due to low-visibility constraints. The acoustic camera, also known as high-resolution forward-looking sonar (FLS), has demonstrated considerable potential in the autonomous monitoring of marine debris, as they are unaffected by water turbidity and dark environments. The appearance of targets in sonar images changes with variations in the imaging viewpoint, while challenges such as low signal-to-noise ratio, weak textures, and imaging distortions in sonar imagery present significant obstacles to debris monitoring based on prior class labels. This paper proposes an optical flow-based method for marine debris monitoring, aiming to fully utilize the time series information captured by the acoustic camera to enhance the performance of marine debris monitoring without relying on prior category labels of the targets. The proposed method was validated through experiments conducted in a circulating water tank, demonstrating its feasibility and robustness. This approach holds promise for providing novel insights into the spatial and temporal distribution of debris.

## I. Introduction

Marine pollution, especially the increase in marine debris, has become a major challenge facing the global ecological environment [1]. Previous studies have shown that marine debris particularly plastic debris, poses one of the most significant environmental threats to ocean ecosystems [2], [3]. The current methods for marine debris removal primarily rely on manual trawling and traditional salvage techniques, which are generally limited to debris near the water surface, as shown in Fig. 1(a). However, in low-visibility marine environments, such as turbid and dark waters, the effectiveness of these conventional methods is significantly constrained. Due to limited visibility, optical cameras face significant challenges in effectively monitoring debris, resulting in delayed removal of debris in deeper waters. Consequently, dominant optical perception techniques are unable to meet the needs of debris monitoring in low-visibility underwater environments, thus limiting the effective cleaning of marine debris.

Different from conventional optical cameras, the acoustic camera [4], [5] is a two-dimensional (2D) forward-looking sonar based on the principle of acoustic imaging, which can obtain high-resolution and high-frame-rate sonar images in low-visibility conditions. Common acoustic cameras include Dual-frequency Identification Sonar (DIDSON) and Adaptive Resolution Imaging Sonar (ARIS), both of which are widely used in underwater imaging and target detection. In addition, the acoustic camera is compact in size and can be flexibly mounted on marine robots for surveys, as shown in Fig. 1(b).

Valdenegro et al. pioneered the use of acoustic cameras for the detection of marine debris, achieving preliminary results in recognition [6] and segmentation [7]. Their work is highly innovative, as it demonstrated the effectiveness of acoustic cameras in debris detection, despite the absence of dynamic monitoring. Additionally, they incorporated deep learning techniques to enhance the accuracy of debris perception.

The deep learning-based debris monitoring methods have shown promising results in detecting static seabed debris, particularly for targets with fixed shapes and stable positions, where high accuracy can be achieved. However, these methods typically rely on large amounts of labeled categories for training, which presents significant challenges in the monitoring of marine debris, especially microplastic waste. First, microplastics lack uniform visual representations [8], and their diverse forms increase the difficulty in data collection. Second, even for clearly defined types of debris, there is often morphological diversity, such as flattened bottles or broken bags, and these deformed targets may deviate from the raw class features. Therefore, marine debris monitoring methods based on prior category labels may not fully leverage their advantages when dealing with these heterogeneous targets, thereby impacting the efficiency of debris removal.

Therefore, it is crucial to propose a more robust method for marine debris monitoring to address the limitations of existing approaches [9]. The method should be adaptable to various types and forms of debris and capable of effectively handling the challenges posed by low visibility and dynamic environments. Finally, efficient marine debris monitoring can provide objective data support and a scientific basis for marine debris removal under low-visibility conditions, thereby offering a reliable foundation for related decision-making.

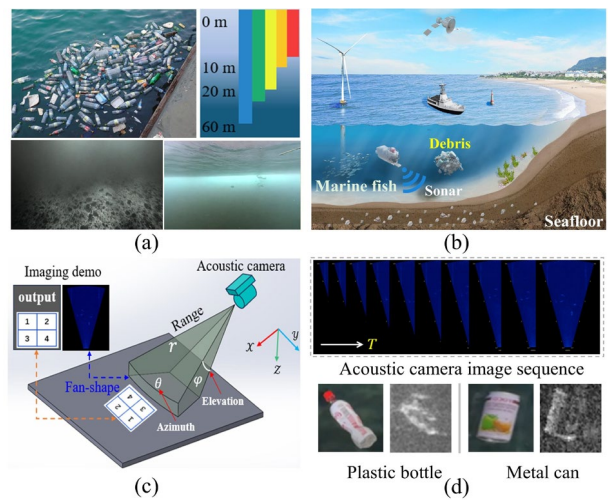

Fig. 1. Application of acoustic cameras in marine debris monitoring.

## II. MOTIVATION

*1) Full utilization of existing sensor information:* Acoustic cameras are high-frame-rate imaging sonar that can capture fine time-series information. However, the dynamic visual cues of sonar targets are usually not fully exploited.

*2) Monitoring dynamic marine debris:* Debris is not only distributed on the static seabed but may also float in the water or move with the ocean currents. Traditional monitoring methods based on static features are prone to failure in dynamic scenes, while optical flow can capture the movement trajectory of debris, providing important support for the detection and tracking of dynamic debris.

*3) Eliminating the dependence on prior category labels:* The optical flow is mainly based on the motion characteristics of the target rather than the appearance characteristics. This enables it to effectively perceive and monitor marine debris when the category labels are insufficient or the target appearance characteristics are diverse, such as entangled debris and microplastics.

*4) Addressing the issue of similar visual representation of targets in sonar imagery:* In acoustic camera images, the visual features of underwater targets are often ambiguous. Fish and certain types of marine debris, such as plastic bottles and metal cans, may exhibit similar shapes or reflective properties, posing significant challenges to target detection methods based on static images. Optical flow, by capturing the motion patterns of targets across consecutive frames, provides additional discriminative information. Fish typically exhibit distinct biological motion characteristics, such as rhythmic tail movements or sharp directional changes, while the movement of marine debris is primarily influenced by environmental factors, such as currents or buoyancy, resulting in relatively stable or random trajectories. These differences in motion patterns make optical flow a powerful tool for distinguishing visually similar targets in sonar images, offering technical support for the robust monitoring of underwater targets.

*5) Revealing the spatial and temporal distribution characteristics of marine debris:* The introduction of acoustic cameras and optical flow technology for marine debris monitoring allows for precise tracking of target trajectories and reveals their dynamic changes over time. By deploying acoustic cameras for long-term underwater observation, key information such as accumulation areas, drift direction, and speed of debris can be effectively identified. This functionality provides objective and reliable data support for the planning and execution of marine debris cleanup tasks.

## III. METHODOLOGY

### A. Principle of acoustic camera imaging

The acoustic camera is an active sonar device that detects targets forward by emitting a wide range of three-dimensional (3D) fan-shaped sound waves and receiving echo signals reflected from the surface of the object for imaging [10], [11], as displayed in Fig. 1(c). The 3D points within the sonar field of view (FOV) can be represented as $(r, \theta, \varphi)$ in polar coordinates and converted into Euclidean coordinates by Equation (1). The $r$ represents range, $\theta$ refers to the azimuth angle, and $\varphi$ denotes the elevation angle.

$$[E_X \; E_Y \; E_Z]^T = [r\cos\varphi\cos\theta \quad r\cos\varphi\sin\theta \quad r\sin\varphi]^T \quad (1)$$

The visual data captured by the acoustic camera exhibits continuity and is stored in a video-like format, as displayed in Fig. 1(d). Furthermore, users have the capability to manually adjust the imaging frame rate to meet specific application requirements or optimize image quality. Therefore, high-quality time series data can be captured as the distance $\mathcal{D}$ from the target to the acoustic camera changes, as shown in Fig. 2.

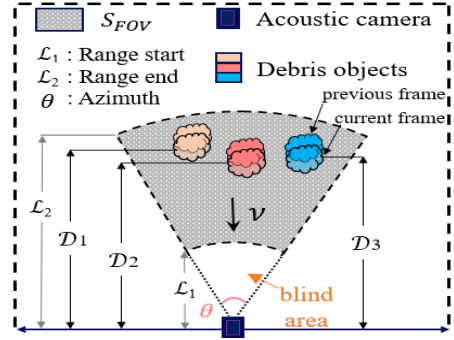

Fig. 2. The illustration of capturing time series data on marine debris.

### B. Methodology framework

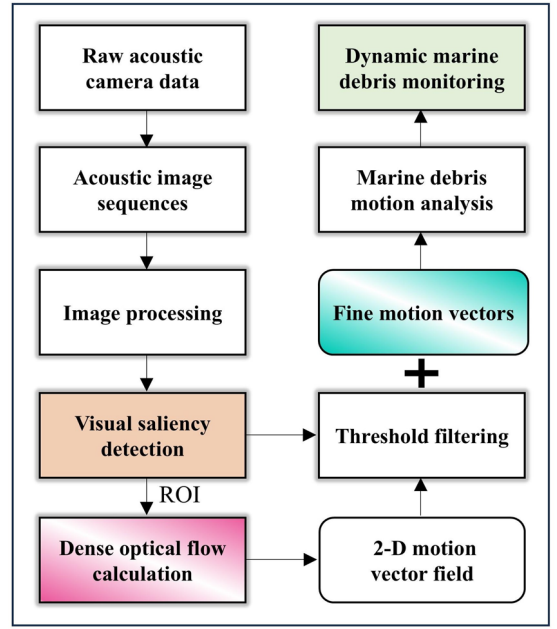

Fig. 3. The pipeline of our proposed method for marine debris monitoring.

This subsection presents the integrated pipeline for marine debris monitoring based on acoustic camera time-series data and optical flow analysis, as illustrated in Fig. 3. First, the raw data collected by the acoustic camera is converted into an image sequence, with frames containing no targets manually removed to ensure data validity. Then, a series of image preprocessing operations are applied to enhance the quality of the input data, providing more accurate information for subsequent analysis. Visual saliency detection is introduced to extract the region of interest (ROI), laying the foundation for target monitoring. Next, optical flow algorithms are applied to estimate the motion vectors of targets in sonar images, capturing the movement information of the targets. Following this, post-processing techniques are used to filter the optical flow results, yielding refined motion vectors. Finally, based on these fine motion vectors, the movement trajectories of debris targets are constructed, enabling dynamic monitoring of the targets in low-visibility underwater environments.

## C. Method feasibility verification and analysis

Dynamic target monitoring is essentially a tracking problem within the field of computer vision, aiming to track specific targets in video or image sequences in real-time. This task involves several key steps, such as target detection, feature extraction and matching, and motion estimation. In the rapid tracking of optical targets, feature matching-based pipelines are widely applied. However, due to the inherent characteristics of sonar images, existing methods face certain limitations in their application to acoustic imaging, primarily regarding the significant challenges in extracting high-quality and reproducible local feature points from sonar images. To visually demonstrate this sonar image processing issue, a comparative experiment was conducted. Specifically, several commonly used feature detection operators, including SIFT [12], FAST [13], ORB [14], and AKAZE [15], [16], were applied to extract local feature points around the targets from optical and sonar image sequences. The positions of these feature points were then visualized. The experimental results are presented in Figs. 4 to 7. All feature detection operators were implemented using OpenCV [17] with default parameter settings to facilitate community reference.

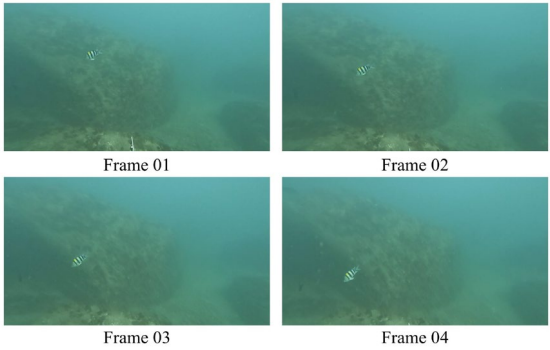

Fig. 4. Sequence of underwater optical camera images of a marine fish.

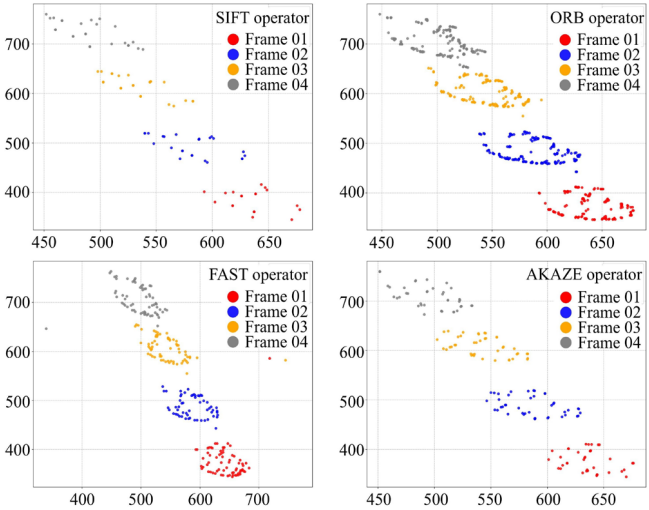

Fig. 5. Extracted feature point positions of various operators on optical images.

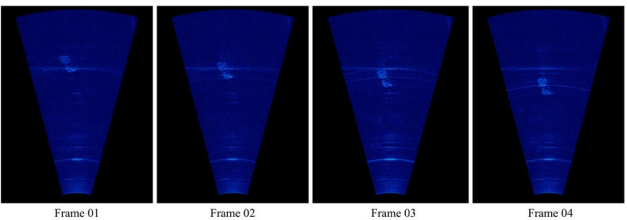

Fig. 6. Sequence of underwater acoustic camera images of a plastic bottle.

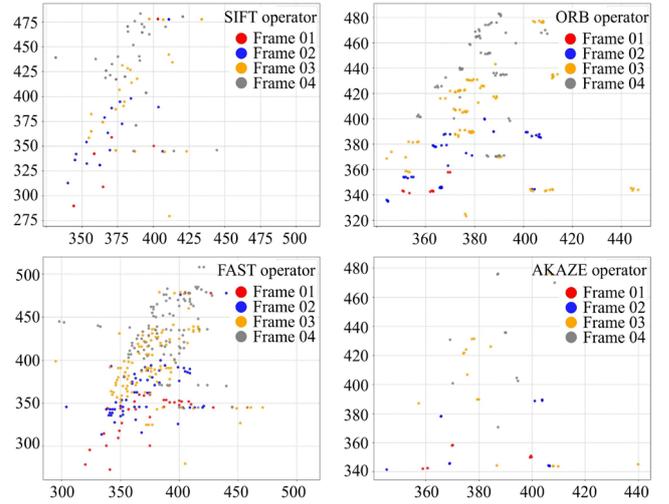

Fig. 7. Extracted feature point positions of various operators on sonar images.

Analysis of the statistical results in Figs. 5 and 7 reveals that, in optical image sequences, the extracted feature points exhibit a regular pattern as the target moves. The positions of the feature points extracted by different algorithms remain relatively stable and follow a consistent pattern with the target's movement. This indicates that feature points in optical images are highly stable and effectively reflect the target's trajectory. In contrast, in sonar image sequences, as the target displaces, the extracted feature points display randomness and dispersion, lacking clear regularity. This phenomenon is likely associated with the low contrast, noise interference, and blurriness of sonar images, which prevents the consistency of feature points across frames. The results of this test highlight the significant challenges faced by traditional feature point extraction methods in sonar images, which fail to provide stable and repeatable feature points. This suggests that a new marine debris monitoring method needs to be developed to improve its robustness and accuracy in complex sonar images.

## D. Acoustic camera imagery processing

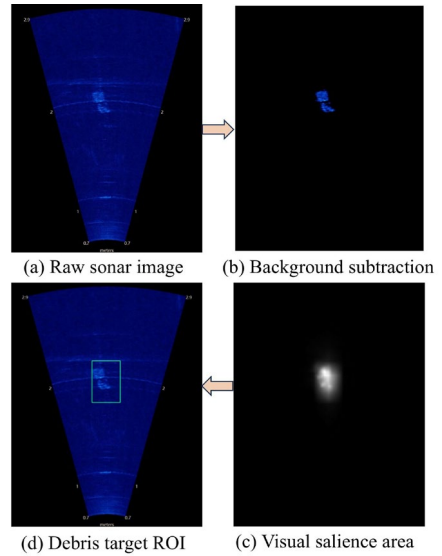

Fig. 8. ROI generation based on visual saliency detection.

Generating ROIs based on saliency detection prior to computing optical flow vectors in sonar images is a pragmatic decision. First, the majority of sonar image content typically consists of water or seabed background unrelated to debris. These regions will significantly increase computational redundancy and introduce irrelevant information interference. By employing visual saliency detection [18] to generate ROIs, the focus is effectively directed toward regions where debris targets are likely to exist, reducing the influence of irrelevant backgrounds and enhancing the efficiency and accuracy of optical flow computation. Second, due to the low signal-to-noise ratio (SNR) and imaging distortions characteristic of sonar images, directly computing optical flow across the entire image may result in degraded algorithm performance, such as the generation of spurious vectors or false detections. Restricting optical flow computation to salient regions significantly reduces noise interference, improves the robustness and precision of optical flow estimation, and better supports subsequent marine debris monitoring. The process of generating a debris target ROI based on visual saliency detection is shown in Fig. 8, and the background subtraction processing is implemented based on ARIScope software [5].

Additionally, in acoustic camera images, the appearance of marine debris targets is often not fully continuous, with void regions or noise, which will directly affect the accuracy of subsequent optical flow calculations. This is particularly problematic for debris such as bottles and cans, whose surfaces should be continuous. However, discontinuities in the sonar image may lead to errors in optical flow estimation, thereby impacting debris monitoring performance.

To address this issue, this study proposes using image inpainting techniques to process the acoustic camera images. Image inpainting can effectively improve the accuracy of optical flow calculation by filling gaps and restoring the continuity of the target surface, thereby providing more accurate and stable dynamic data support for marine debris monitoring. This study uses three image inpainting methods implemented in OpenCV, including INPAINT_TELEA, INPAINT_NS algorithms, and the multi-scale inpainting method based on the inpaint_biharmonic algorithm [17]. These methods effectively fill void regions in sonar images, restoring the continuity of debris target surfaces, thereby significantly improving the accuracy and stability of subsequent optical flow calculations. During the sonar image inpainting process, all algorithms are applied with default parameter settings to ensure reproducibility and facilitate comparative analysis in future studies. Moreover, guided filtering [19] is introduced to smooth the inpainting results further, enhancing the overall quality of the inpainted images. An inpainting demo of a debris image is displayed in Fig. 9.

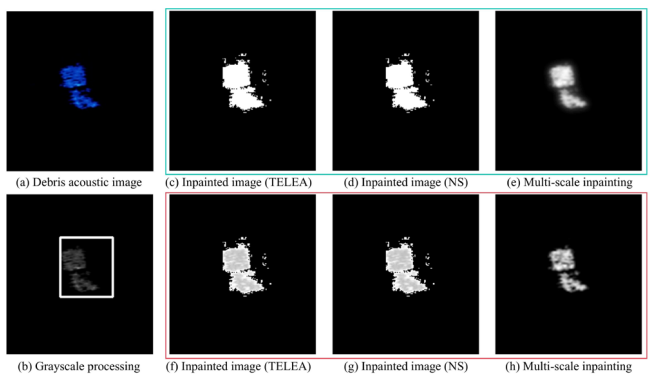

Fig. 9. Inpainted debris target using various algorithms. The content within the green box displays the results processed by guided filtering.

### E. Optical flow calculation model

Given the inherent challenges of acoustic camera images in practical applications, such as weak texture, low contrast between targets and background, and significant noise interference, traditional sparse optical flow methods often struggle to extract sufficient and repeatable local feature points [20], resulting in unreliable motion vector estimation outcomes. Furthermore, although some deep learning-based pipelines show strong capabilities in optical flow estimation, their high dependence on computational resources and the difficulty of acquiring training data [21], [22], particularly in dynamic marine debris monitoring applications, limit their feasibility in real-time processing scenarios.

Considering these challenges, this study introduces the Farneback algorithm [23] for optical flow estimation. As a dense optical flow method, the Farneback algorithm's advantage lies in its independence from local feature point extraction, relying instead on pixel-level grayscale variations and the assumption of flow smoothness for estimation. This makes it robust for motion estimation tasks in complex sonar imaging environments. An example of optical flow estimation for marine debris is shown in Fig. 10.

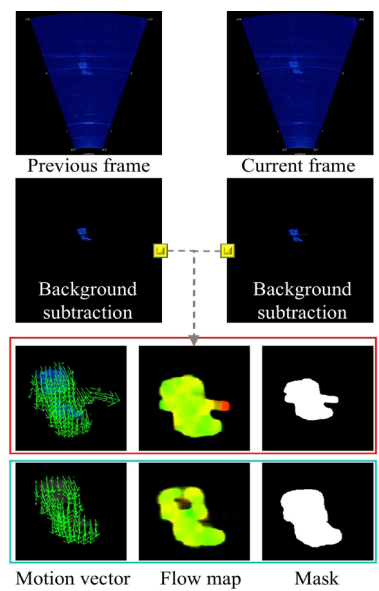

Fig. 10. The optical flow calculation results of a plastic bottle. The content within the green box displays the results after image inpainting.

### F. Approach precision verification

To validate the effectiveness of the proposed method, this study uses a video-like dataset acquired by an acoustic camera and manually annotated real motion trajectories of targets. Specifically, we manually draw the bounding box around the target in each frame and record the center position of the target for each frame. The motion vectors calculated from these data serve as the ground truth (GT), and the overall motion trajectory reflects the true motion path of the debris target, as displayed in Fig. 11. Next, we compare the motion vectors estimated by the proposed method with the GT.

Furthermore, this study conducts a qualitative assessment by visualizing the motion trajectory of debris targets, followed by a quantitative analysis through the calculation of the motion speed of debris targets. The method for calculating the target's motion speed is given by Equation (2).

$$V = \Delta L \times R \times F \qquad (2)$$

where $V$ denotes the inferred speed of the debris target, $\Delta L$ represents the pixel displacement of the target center between adjacent frames calculated using optical flow, $R$ represents the spatial resolution of the acoustic camera image, and $F$ is the set imaging frame rate of the acoustic camera.

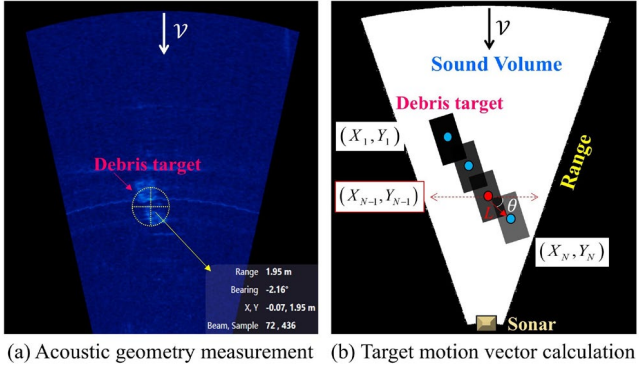

(a) Acoustic geometry measurement  (b) Target motion vector calculation

Fig. 11. Calculation of the real motion vectors (GT) of the debris target.

## IV. EXPERIMENTAL SETTINGS

### A. Overview

This study designed and conducted two independent tank experiments to systematically simulate the dynamic behavior of marine debris in natural marine environments, focusing on horizontal drift motion (along the water flow direction) and vertical movement (along the depth direction). The acoustic camera used in these experiments was an ARIS EXPLORER 3000 with an identification frequency of 3.0 MHz. In the process of debris monitoring, the frame rate of the acoustic camera is set to 10 frames per second (FPS), and the resolution of acoustic camera imaging is 2.9mm.

### B. Horizontal monitoring experiment

This experiment was conducted in a circulating water tank, and the testing site is shown in Fig. 12(a). Debris targets were placed on the water surface and moved with the current, and the flow velocity can be controlled. The sonar image of plastic bottles captured by the acoustic camera is shown in Fig. 12(b). Fig. 12(c) provides relevant parameters of the experiment site.

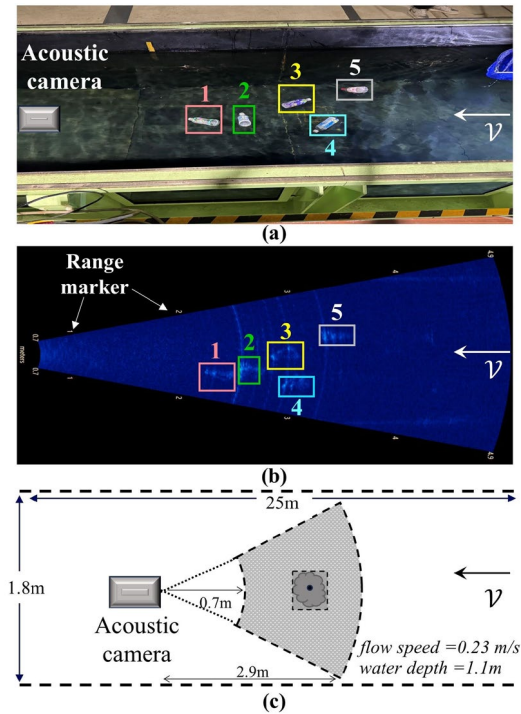

Fig. 12. Experimental site and acoustic camera imaging demonstration.

### C. Vertical monitoring experiment

To evaluate the sonar's capability to monitor the vertical movement of marine debris, this experiment utilized a method in which debris targets were fixed to poles. External water flow forces of varying intensities were applied to induce the motion of the debris targets along the depth axis. The experimental setup is depicted in Fig. 13. To enhance the accuracy of the vertical movement observation, the acoustic camera was rotated 90 degrees around the roll axis, aligning its monitoring view with the direction of target movement. This experimental configuration enabled the acoustic camera to successfully capture the movement data of debris targets along the depth axis, as illustrated in Fig. 14.

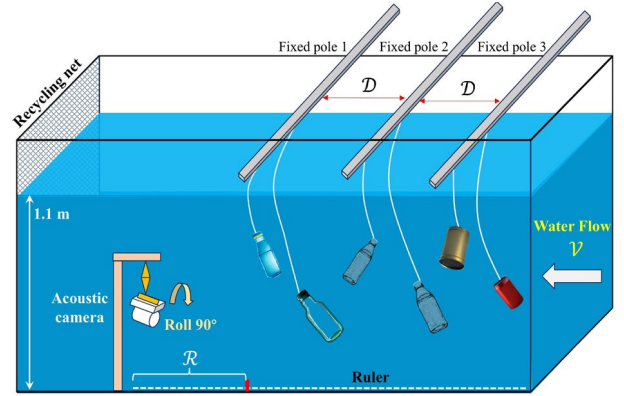

Fig. 13. Experiment on vertical movement monitoring of marine debris.

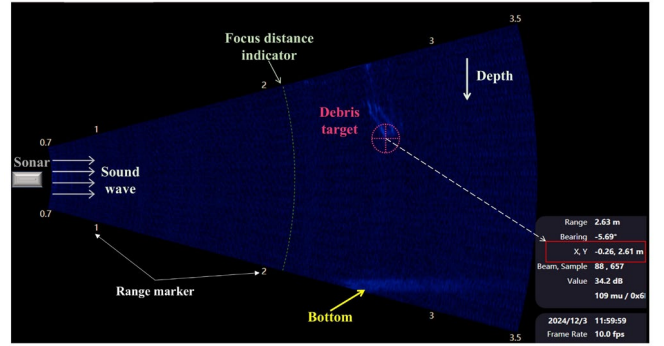

Fig. 14. Detailed vertical movement monitoring using an acoustic camera.

## V. RESULTS AND EVALUATION

### A. Evaluation of horizontal motion monitoring

In the horizontal monitoring verification experiment, this study selected a plastic bottle target and a plastic bag target for testing. The plastic bottle, as a representative of structured debris, exhibits relatively stable visual representations in acoustic camera images. In contrast, the plastic bag is typically considered unstructured debris, as its visual representation in sonar images changes continuously with movement. The motion trajectories estimated using the proposed method are shown in Figs. 15 and 16, where the yellow trajectory line on the left represents the manually labeled true trajectory (GT), and the green trajectory line on the right represents the results estimated by the proposed approach.

The results above indicate that the method proposed in this study provides relatively accurate monitoring of plastic bottles, effectively capturing their motion trajectories. In contrast, the monitoring of plastic bags shows some deviation, likely due to the continuous shape changes of the plastic bags during movement, which may lead to some optical flow estimation errors. However, despite these deviations, the overall motion trajectory region remains consistent with the GT, suggesting that the proposed method is effective in reflecting the motion trajectory of plastic bags.

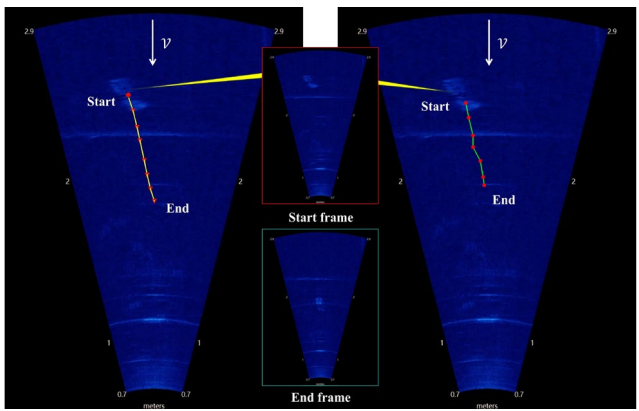

Fig. 15. Visualization of the estimated horizontal motion trajectory (bottle).

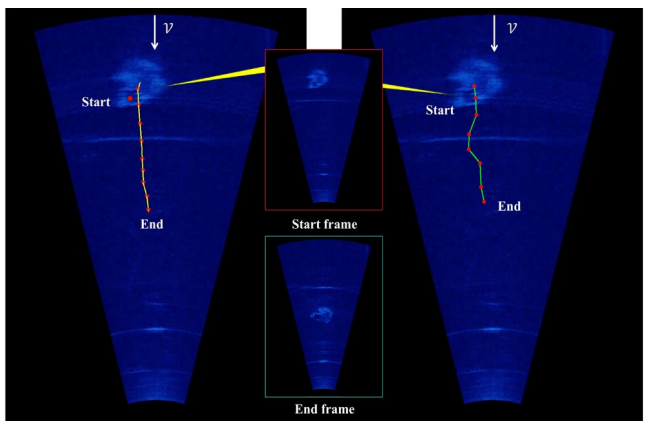

Fig. 16. Visualization of the estimated horizontal motion trajectory (bag).

Additionally, considering that the motion of debris targets in the horizontal monitoring experiment is driven by the force of the water flow, we estimated the targets' motion speed and compared it with the real water flow velocity to quantitatively evaluate the accuracy of the proposed method. Specifically, the motion speed of 10 targets was selected for calculation, and their average value was taken as the final motion speed. The relevant statistical results are shown in Table I.

TABLE I. THE ESTIMATED MOTION SPEED OF VARIOUS DEBRIS

| Debris target | Estimated Motion speed | Real Flow velocity |
|---|---|---|
| Plastic bottles | 0.24 m/s | 0.23 m/s |
| Plastic bags | 0.20 m/s | 0.23 m/s |

It is evident that the target motion speed estimated by the proposed method aligns closely with the water flow speed, thereby validating the effectiveness of the approach. Notably, the motion speed estimation for plastic bottles exhibits higher accuracy compared to that for plastic bags.

### B. Evaluation of vertical motion monitoring

In the vertical motion monitoring experiment, the same target setup as in the horizontal monitoring experiment was used, and the motion trajectory of the debris in the depth direction was visualized, as shown in Figs. 17 and 18. Similar to the horizontal monitoring results, the proposed method displayed higher accuracy when applied to plastic bottles, while some errors were observed in the application to bags.

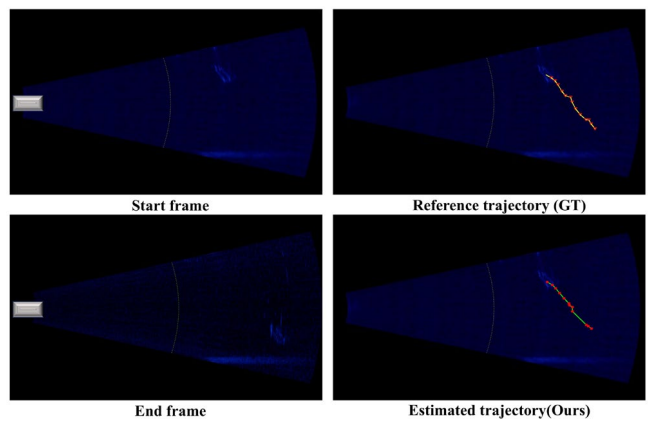

Fig. 17. Visualization of the estimated vertical motion trajectory (bottle).

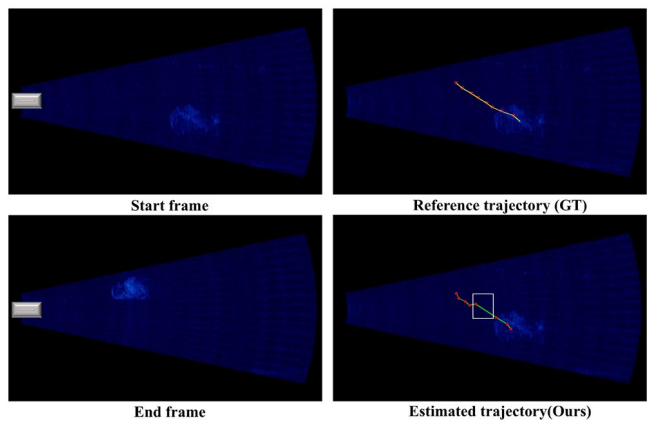

Fig. 18. Visualization of the estimated vertical motion trajectory (bag).

### C. Analysis of trajectory estimation error

During the process of estimating the motion trajectory of plastic bags, obvious deviations were observed in white box regions, as shown in Fig. 18. A visual analysis of the optical flow calculation process revealed that in areas with large errors, the moving target was incorrectly identified as two separate targets, and there was a significant deviation in the estimation of the target's centroid, as display in Fig. 19.

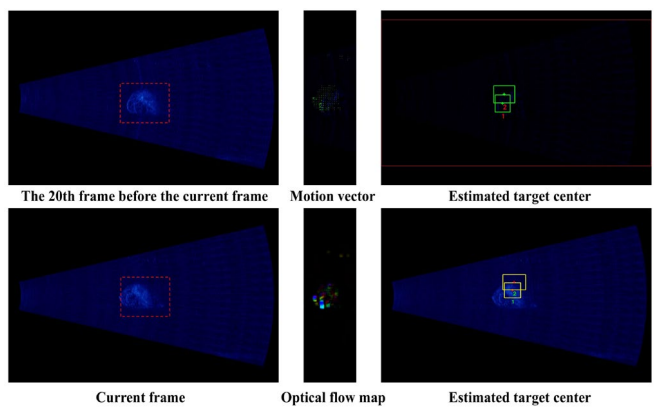

Fig. 19. Visualization of motion vector analysis.

This phenomenon is closely related to the choice of frame interval in optical flow estimation. When the frame interval is large, the accuracy of the optical flow estimation decreases. This issue arises from the inherent characteristics of acoustic imaging, where the imaging of the target changes significantly as the relative position between the target and the sonar varies. Therefore, selecting a suitable frame interval is crucial for balancing the accuracy and efficiency of the proposed method.

## D. Multiple marine debris monitoring testing

This paper presents the monitoring results of the proposed method on multiple marine debris targets, as shown in Fig. 20. The results indicate that our proposal demonstrates good estimation accuracy and robustness in multi-target scenes. This data represents plastic bags captured in the same water tank under conditions of increased flow velocity (0.59 m/s).

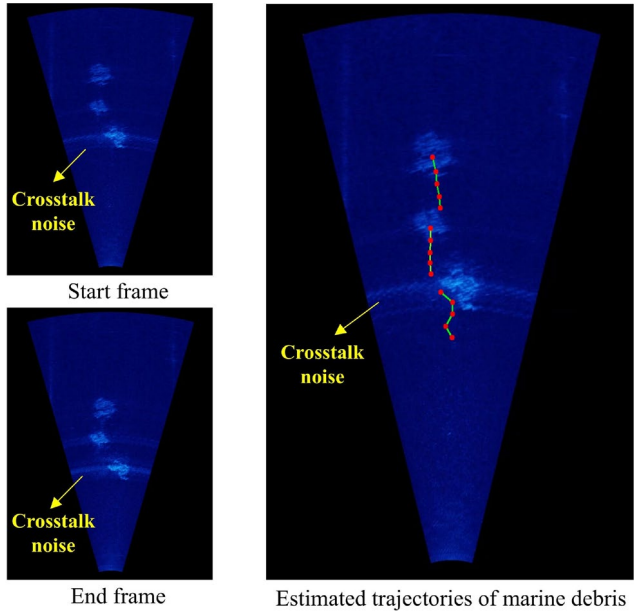

Fig. 20. Estimated movement trajectories of multiple marine debris.

In addition, Fig. 20 shows that there is some deviation in the trajectory estimation of the bottom target. To further analyze this issue, the intermediate process of the trajectory estimation is visualized, as shown in Fig. 21. The results indicate that this error mainly originates from the influence of crosstalk noise. Although background subtraction operations were taken during the preliminary sonar image preprocessing, crosstalk noise was not completely eliminated.

Actually, the crosstalk noise is a persistent type of noise, typically distributed on both sides of the target and exhibiting motion vector characteristics similar to the target. Therefore, during optical flow calculation, crosstalk noise is easily misidentified as part of the target. This phenomenon highlights the urgent need for developing a more effective crosstalk noise removal algorithm.

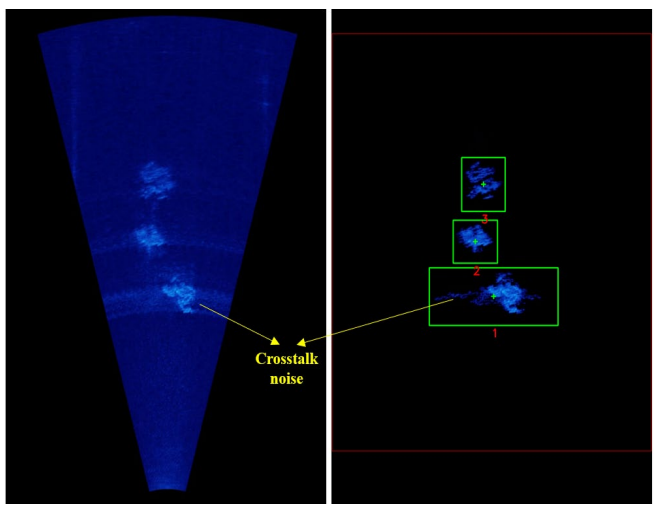

Fig. 21. The impact of crosstalk noise on target trajectory estimation.

## VI. DISCUSSION

This study proposes an approach that integrates acoustic cameras and optical flow to monitor the motion information of marine debris, including the velocity and trajectory of targets. In low-visibility environments, the method effectively reveals the spatial distribution characteristics and dynamic behaviors of marine debris. By implementing preprocessing steps such as background subtraction and image inpainting, the approach mitigates the prevalent issues of low SNR and texture voids in sonar images, thereby enhancing the accuracy of optical flow calculations. Experimental results demonstrate that the method can accurately monitor debris motion in both horizontal and vertical directions, showing high reliability in estimating the velocity and trajectory of structured targets such as plastic bottles. However, for unstructured targets with significant deformation, such as plastic bags, trajectory estimation still exhibits certain deviations, indicating that further optimization is needed to improve the adaptability. Some of the planned enhancement directions are as follows:

*1) Performance testing in dynamic environments:* To comprehensively evaluate the algorithm's performance in real marine environments, field experiments will be conducted under complex dynamic conditions, such as high-intensity water flow and multi-target interference. Through systematic analysis, the robustness and adaptability of the proposed method in practical application scenes will be validated.

*2) More comprehensive identification of marine debris:* By integrating lightweight deep learning models, a smart system capable of real-time marine debris recognition will be developed. Motion vectors will be incorporated as a novel feature dimension to enhance target recognition accuracy. Particularly in distinguishing highly similar targets, such as fish and debris, motion features can serve as crucial auxiliary information, improving recognition performance.

*3) Assisting marine robots in debris collection:* The proposed method will be deeply integrated with the control system of intelligent marine robots to guide precise localization and collection of marine debris in dynamic environments. The focus of the research is on how to achieve efficient waste collection and path planning optimization under complex marine conditions, thus advancing marine cleaning technology toward more intelligent solutions.

## VII. CONCLUSION

This study presents a novel method for monitoring marine debris in low-visibility environments, leveraging acoustic camera time-series data and optical flow techniques. This approach includes the extraction of time series information, sonar image preprocessing, optical flow calculation, and motion vector estimation, enabling the precise capture of the movement characteristics of marine debris targets and revealing their spatiotemporal distribution patterns. The proposed method offers notable advantages, including ease of operation and low cost, making it well-suited for marine environmental monitoring and providing scientific data to support the removal and management of marine debris. Furthermore, the method can be applied to path planning and target classification enhancement in marine robots-based debris collection processes. In the future, we will further evaluate the robustness of the proposed method in more complex field environments and validate it using different types of acoustic imaging devices, such as DIDSON, to explore the generalization ability of the methodology.

ACKNOWLEDGMENT

The authors gratefully acknowledge the cooperation of Windy Network Inc. in the preparation of the experiments. The authors also thank Mr. Dianhan Xi, Mr. Yilong Zhang, Mr. Kaede Furuichi, Mr. Weizhen Liang, Ms. Zhiyan Ren, and Baoxi Huang, who are students at the Graduate School of Frontier Sciences, The University of Tokyo, and Mr. Yoshida, the technical manager of the water tank at the Institute of Industrial Science, The University of Tokyo, for their assistance in conducting the experiments.